%% file: cas-sc-template.tex
\documentclass[a4paper,fleqn]{cas-dc}

\usepackage{cite}
\usepackage{amsfonts,amssymb,booktabs}
\usepackage{amsmath}
\usepackage{algorithm}
\usepackage{algorithmic}
\usepackage{array,multirow}
\usepackage[caption=false,font=normalsize,labelfont=sf,textfont=sf]{subfig}
\usepackage{fixltx2e}
\usepackage{stfloats}
\usepackage{url}
\usepackage{setspace}

\usepackage{amssymb}
\usepackage[switch]{lineno}     
\usepackage{booktabs}           
\usepackage{multirow}           
\usepackage{array}              
\usepackage{threeparttable}     
\usepackage{color}              
\usepackage{float}              
\usepackage{makecell}           
\usepackage{algorithm}          
\usepackage[numbers,sort&compress]{natbib}
\usepackage[justification=centering]{caption}		
\usepackage{array} 
\usepackage{units} 

\def\tsc#1{\csdef{#1}{\textsc{\lowercase{#1}}\xspace}}
\tsc{WGM}
\tsc{QE}

\begin{document}
\let\WriteBookmarks\relax
\def\floatpagepagefraction{1}
\def\textpagefraction{.001}

\shorttitle{}    

\shortauthors{}  

\title [mode = title]{Efficient Visual Fault Detection for Freight Train Braking System via Heterogeneous Self Distillation in the Wild}  

\tnotemark[1] 

\tnotetext[1]{Corresponding author.}

%


\author[1,2,3]{Yang Zhang}[]
\author[1,2]{Huilin Pan}[]
\author[1,2]{Yang Zhou}[]
\author[1,2]{Mingying Li}[]
\author[1,2]{Guodong Sun}[]
\cormark[1]






\affiliation[1]{organization={School of Mechanical Engineering},
            addressline={Hubei University of Technology}, 
            city={Wuhan},
            postcode={430068}, 
            country={China}}
\affiliation[2]{organization={Hubei Key Laboratory of Modern Manufacturing Quality Engineering},
            addressline={Hubei University of Technology}, 
            city={Wuhan},
            postcode={430068}, 
            country={China}}
\affiliation[3]{organization={National Key Laboratory for Novel Software Technology},
			addressline={Nanjing University}, 
			city={Nanjing},
			postcode={210023}, 
			country={China}}



\begin{abstract}
Efficient visual fault detection of freight trains is a critical part of ensuring the safe operation of railways under the restricted hardware environment. Although deep learning-based approaches have excelled in object detection, the efficiency of freight train fault detection is still insufficient to apply in real-world engineering. This paper proposes a heterogeneous self-distillation framework to ensure detection accuracy and speed while satisfying low resource requirements. {The privileged information in the output feature knowledge can be transferred from the teacher to the student model through distillation to boost performance.} We first adopt a lightweight backbone to extract features and generate a new heterogeneous knowledge neck. Such neck models positional information and long-range dependencies among channels through parallel encoding to optimize feature extraction capabilities. Then, we utilize the general distribution to obtain more credible and accurate bounding box estimates. Finally, we employ a novel loss function that makes the network easily concentrate on values near the label to improve learning efficiency. Experiments on four fault datasets reveal that our framework can achieve over 37 frames per second and maintain the highest accuracy of 98.88$\%$ in comparison with traditional distillation approaches. Moreover, compared to state-of-the-art methods, our framework demonstrates more competitive performance with lower memory usage and the smallest model size.
\end{abstract}



\maketitle

\input{files/1-Introduction}

\input{files/2-Related_works}
\input{files/3-Methods}
\input{files/4-Experiments}

\input{files/5-Conclusion}

\printcredits

\bibliographystyle{cas-model2-names}

\bibliography{refs}

\bio{}
\endbio

\end{document}

%% file: files/1-Introduction.tex
\section{Introduction}
\label{Introduction}
The vehicle braking system is an essential component for guaranteeing the safe and efficient operation of the freight train. The performance of the braking system is inevitably degraded or even functional failure as operating mileage, service time, and continuing influence of the operating environment. Many crucial components in the braking system must be carefully inspected, such as the bogie block key, cut-out cock, dust collector, and fastening bolts on the brake beam~\cite{Zhang_TIM}. The removal or displacement of these parts will have a significant impact on driving safety. Moreover, image acquisition devices are usually located in the wild, and the complexity of the environment causes occlusion and ambiguity in the fault area, which makes fault detection more challenging. Specifically, the wild environment requires fault detectors containing a smaller model size and lower computational complexity. In addition, it is difficult to obtain sufficient features because individual components are usually small and easily contaminated, resulting in blurred edges of damaged objects. The texture of these elements is usually comparable to the background, and they usually have a lot of structural information. 

To enhance the reliability of the braking system and avoid accidents, some researchers have explored several fault detectors for freight trains employing machine learning as a detection approach~\cite{Chang_TIE, Sun}. Due to poor portability and limited detection accuracy, the traditional machine learning-based methods are difficult to meet the actual detection need. With the widening range of object detection applications~\cite{PR_label},~\cite{PR_video},~\cite{PR_underwater},~\cite{AEI1} based on deep learning, automated detect inspection of surface~\cite{AEI2, AEI3} is still a challenging task in the automation field. In recent years, researchers have also proposed some methods with excellent performance for detecting faults of the freight trains~\cite{Zhang_TIM, Zhaiyao_Zhang}. However, most of these detectors are tough for fault areas with blurred, occluded, or contaminated parts to precisely find and locate. Meanwhile, these detectors focus on feature prediction at different scales while ignoring the positional information between different channels and long-range dependencies in space. Such ignorance usually results in the loss of certain valuable information between the fault component and background and degrades the performance of the detector.

\begin{figure}[!t]
	\begin{center}
		\centering
		\includegraphics[width=3.3in]{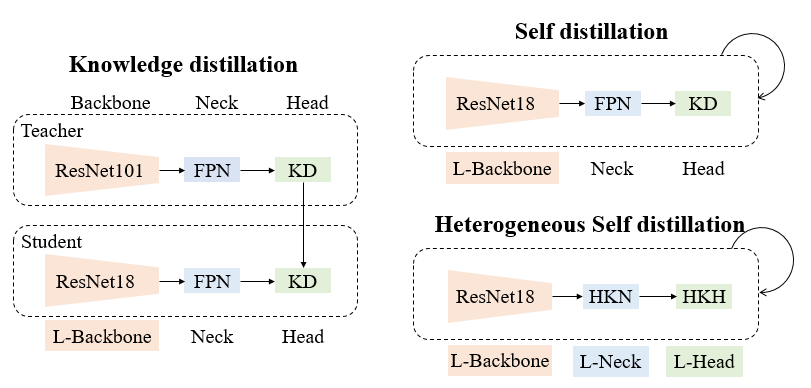}
		\caption{Comparison of traditional knowledge distillation and our proposed heterogeneous self distillation. In the former, the student only uses the lightweight module in the backbone, whereas in the latter, the lightweight module is also utilized by the detection head (\textbf{L-Head}) and neck (\textbf{L-Neck}).}
		\label{fig1}
	\end{center}
\end{figure}

Recently, localization distillation (LD)~\cite{LD_2021} has shown that knowledge distillation is a potential method to improve the localization quality of object detection. Such an approach offers an inspired idea to solve the localization ambiguity for fault edges at the bottom. {As shown in Fig.~\ref{fig1}, traditional knowledge distillation~\cite{LGD, Lad} utilizes the supervision information of the better-performing teacher model to train the student model for better speed and accuracy.} However, most of these algorithms focus on rising the detection performance of students from the perspective of feature imitation, but there is still some potential in extracting localization ambiguity.
To solve the aforementioned issues, we present a heterogeneous self-distillation framework to improve detection efficiency while meeting the constraints of device deployment in the wild. Firstly, we adopt a lightweight backbone to extract features and generate a new heterogeneous knowledge neck to fuse features of diverse scales. In the heterogeneous knowledge neck, we present a feature coordinate attention to help models more precisely locate and identify the region of interest by modeling positional information and long-range dependencies between different channels. Furthermore, we model the localization information as the general distribution to obtain more credible and accurate bounding box estimates. Finally, we adopt a novel loss function to improve learning efficiency.  On four fault datasets, the experimental results show that our framework implements higher accuracy for important braking system components with over 37 frames per second, which fulfills the real-time detection requirement. Furthermore,  we achieve more competitive performance with lower memory usage and the smallest model size in comparison with state-of-the-art detectors.

In short, our contributions are summarized as follows:
\begin{enumerate}[1)]
	\item Aiming at the problem that the spatial correlation between channels being readily overlooked in existing detectors, we explore the significance of positional information and long-range dependencies between different channels for feature extraction of the fault detector in the wild.
	\item We design a lightweight and precise heterogeneous self-distillation framework for real-time fault detection on freight train images under strict resource constraints.
	\item We model the bounding box estimation as general distribution and improve the loss function, enabling the student model to obtain localization information from the teacher more efficiently.
	\item Abundant experiments show that our framework not only meets real-time detection with low hardware resource requirements but also achieves higher accuracy compared with state-of-the-art methods.
\end{enumerate}

The remainder of this paper is structured as follows. Section~\ref{sec:Related Works} discusses recent work about fault detection for freight train images, general object detection, and knowledge distillation. Section~\ref{sec:Framework} introduces the overall structure and each critical module. To evaluate the efficacy of our framework, we performed detailed and abundant experiments in Section~\ref{sec:Experiments}. Finally, the full article is concluded in Section~\ref{Conclusion}.

%% file: files/2-Related_works.tex
\section{Related Works}
\label{sec:Related Works}

\subsection{Fault Detection for Freight Train Images}
{Recent research articles for fault detection of freight train images are listed as follows.}
Chang et al.~\cite{Chang_TIE} reshaped the deformed image to be standard and used three feature enhancement operators to expand the receptive field of the depth feature map, effectively reducing the impact on the accuracy caused by the target size and occlusion. Xu et al.~\cite{AEI4} constructed a causal network graph between the original monitoring feature variables through a causal discovery algorithm, then extracted information features of the braking system of a high-speed train based on the adjacency matrix of the causal network graph to detect the faults. Sun et al.~\cite{Sun} proposed a two-stage hierarchical feature matching model and used the fast adaptive Markov random field (FAMRF) algorithm and exact height function (EHF) to identify the faults in the brake system. However, it lacked the accuracy required due to the complicated process and high computational cost. Unlike traditional methods, deep learning-based methods are more flexible and can handle more complex problems. Zhang et al.~\cite{Zhaiyao_Zhang} proposed a convolutional neural network (CNN)-based fault detector for freight train images (FTI-FDet) including fault region proposal generation and classification. 
To overcome the deployment constraints of detectors in the wild, Zhang et al.~\cite{Zhang_TIM} proposed a lightweight CNN-based fault detector for freight train images (Light FTI-FDet), which significantly decreases the model size while maintaining acceptable levels of accuracy loss. However, these approaches have enormous computing costs that cannot satisfy realistic requirements like real-time and adaptability.

\subsection{General Object Detection}
State-of-the-art detectors generally employ CNN or its variants as the backbone and continuously enhance their performance through various strategies. Recent object detectors often consider two types of pipeline designs: one-stage and two-stage.
Different from one-stage, two-stage detectors~\cite{Cascade_R-CNN, Faster_R-CNN, Dynamic_R-CNN, Grid_R-CNN} typically generated a set of region proposals and classified them to get initial rough predictions and then refined their localizations with detection heads. A typical example is Faster R-CNN~\cite{Faster_R-CNN}, which selects a region proposal network to replace the traditional selective search, then delivers proposal feature maps to the fully connected layer to identify the target category. Subsequently, two-stage object detectors are mostly based on the architecture of Faster R-CNN to improve, such as Cascade R-CNN~\cite{Cascade_R-CNN}, Dynamic R-CNN~\cite{Dynamic_R-CNN}, Grid R-CNN~\cite{Grid_R-CNN}.
In contrast, one-stage detectors~\cite{AutoAssign,CentripetalNet,YOLOX,FoveaBox,Balanced,Sparse_R-CNN} directly predicted the bounding boxes and class labels for fast inference speed. Some anchor-free methods~\cite{CornerNet}, \cite{FCOS}, \cite{Object} further enhances the performance of the detection algorithm by removing inefficient and complex anchors operations. {One-stage detectors could better satisfy the requirements for fault detection since these detectors used fewer parameters and computation than two-stage methods. }
In addition, Lin et al. presented a feature pyramid network (FPN)~\cite{FPN}, which achieves a good prediction effect by fusing feature maps with different levels to realize a fusion of high-resolution and high-semantic information. The attention mechanism has also shown excellent performance in detection tasks. Xu et al.~\cite{PR1} captured the structural consistency of the prediction results by introducing spatial localization information and covariance features. Abraham et al.~\cite{PR2} trained visual attention maps as new data and images to achieve better performance in image classification tasks. 
Although the aforementioned methods have high accuracy, they are still unable to satisfy the model size and detection speed requirements of real-world scenarios.

\subsection{Knowledge Distillation}
{Knowledge distillation (KD) is a method to transfer knowledge from complex to compact models for improving the performance of compact models~\cite{PR_incremental, PR_co-distillation, PR_action}. Recent studies already effectively use KD to raise the performance of the detectors.}
Chen et al.~\cite{ReviewKD} examined the influence of various cross-level connections between teacher and student networks and combined the review mechanism with multi-layer distillation to streamline the distillation procedure.
Nguyen et al.~\cite{Lad} redistributed labels and used the prediction of the teacher network to evaluate the cost of label allocation instead of directly distilling the student model as a soft label. Zhang et al.~\cite{LGD} proposed a self-distillation framework including a sparse label appearance encoder, inter-object relation adapter, and intra-object knowledge mapper. 
Kang et al.~\cite{ICD} encoded information for each instance of a specified condition into a query and devoted more attention between queries and keywords measure correlation.

However, in the real scenario of train fault detection, the computing capability of hardware resources is frequently constrained. Simultaneously, the fault detector is required to perform high-precision real-time detection for safety concerns. The existing distillation method~\cite{LD_2021} has shown strong ability in compact model learning, but there are still some limitations in capturing localization ambiguity. To overcome the above limitations, we investigate the heterogeneous self-distillation framework and use general distribution to obtain more reliable and accurate bounding box estimates. Moreover, we also conduct an in-depth study of the correlation between channels and spaces and capture the corresponding global relationship. 

%% file: files/3-Methods.tex
\section{Framework}
\label{sec:Framework}
In this part, we present a heterogeneous self-distillation for freight train fault detection (HSD FTI-FDet), which consists of a lightweight backbone, heterogeneous knowledge neck (HKN), and heterogeneous knowledge head (HKH), which is shown in Fig.~\ref{fig3}. To catch the spatial information between channels, we design a feature coordinate attention module to model positional information and long-range dependencies. Then, we model the localization information as a general distribution to obtain more reliable and accurate bounding box estimates. Finally, we optimize the loss function to pass the localization information to the student model.
\begin{figure*}[t]
	\centering
	\includegraphics[width=4.5in]{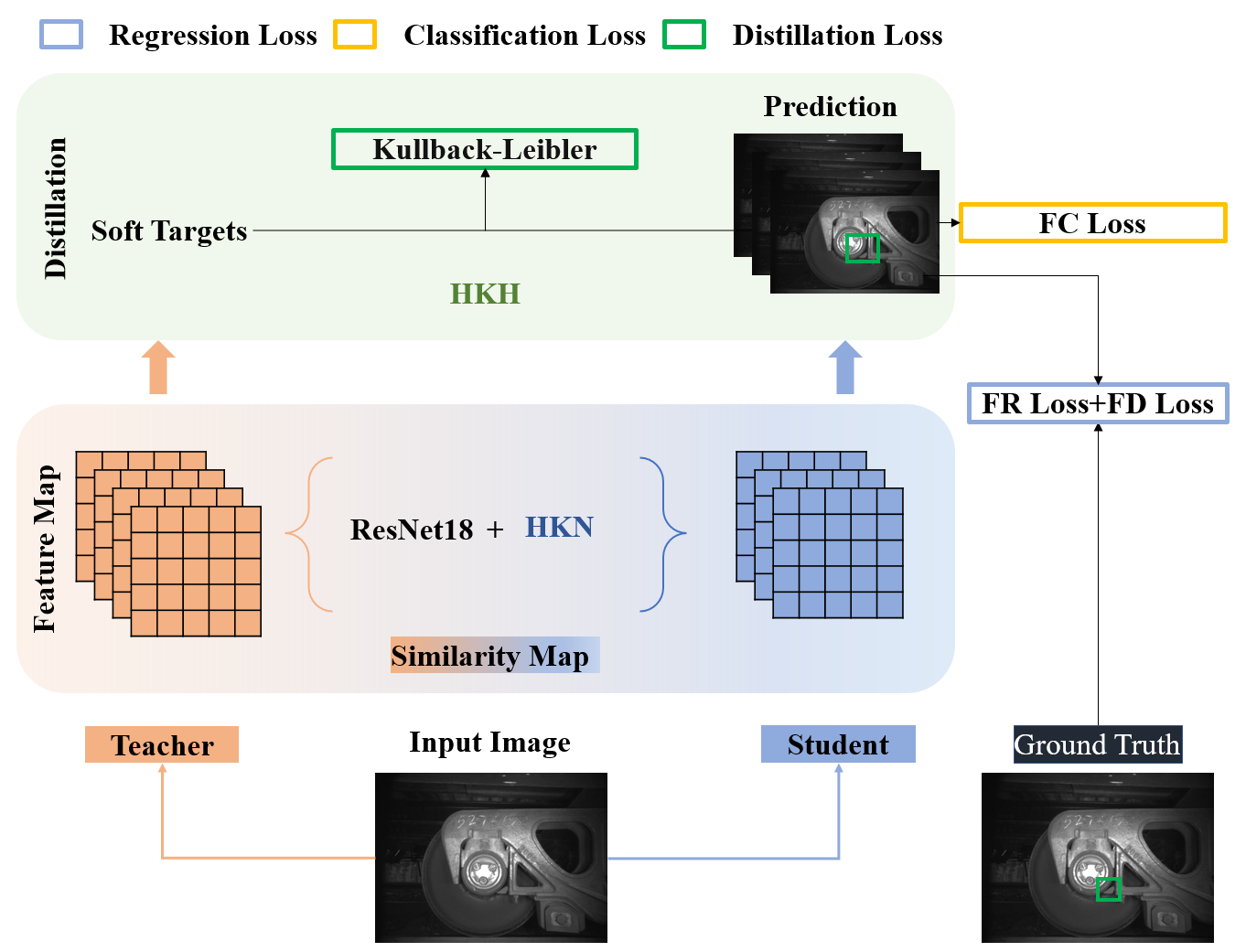}
	\caption{The overall structure of our method. An enhanced module is proposed to model positional information and long-range dependencies. {During the backpropagation process, the regression loss is calculated by fault distribution loss (\textbf{FD Loss}) and fault regression loss (\textbf{FR loss}). The classification loss is computed by fault classification loss (\textbf{FC Loss}), and the distillation is computed by Kullback-Leibler divergence.}}
	\label{fig2}
\end{figure*}

\subsection{Heterogeneous Self Distillation}
\label{HSD}
Since train fault detection generally occurs in the wild, the hardware resources required by the detection model are constrained. Moreover, the compact model and high precision requirements are frequently incompatible with the current detectors. To address this issue, we investigate the applicability of knowledge distillation for train fault detection. By transferring the knowledge from the teacher to the student, distillation can compress the model while allowing the student to learn more generalization capability from the teacher to increase the accuracy. However, when using a complex model as a teacher, the computational cost will rise accordingly, which is not appropriate for a restricted environment of train fault detection. We explore the self-distillation method to extract student models through soft targets and privileged information, omitting the training process of complex teacher models. Nevertheless, the self-distillation model still has a large model size, which cannot meet the requirements for fault detection of freight train braking systems. 

\begin{figure*}[!t]
	\begin{center}
		\centering
		\includegraphics[width=6in]{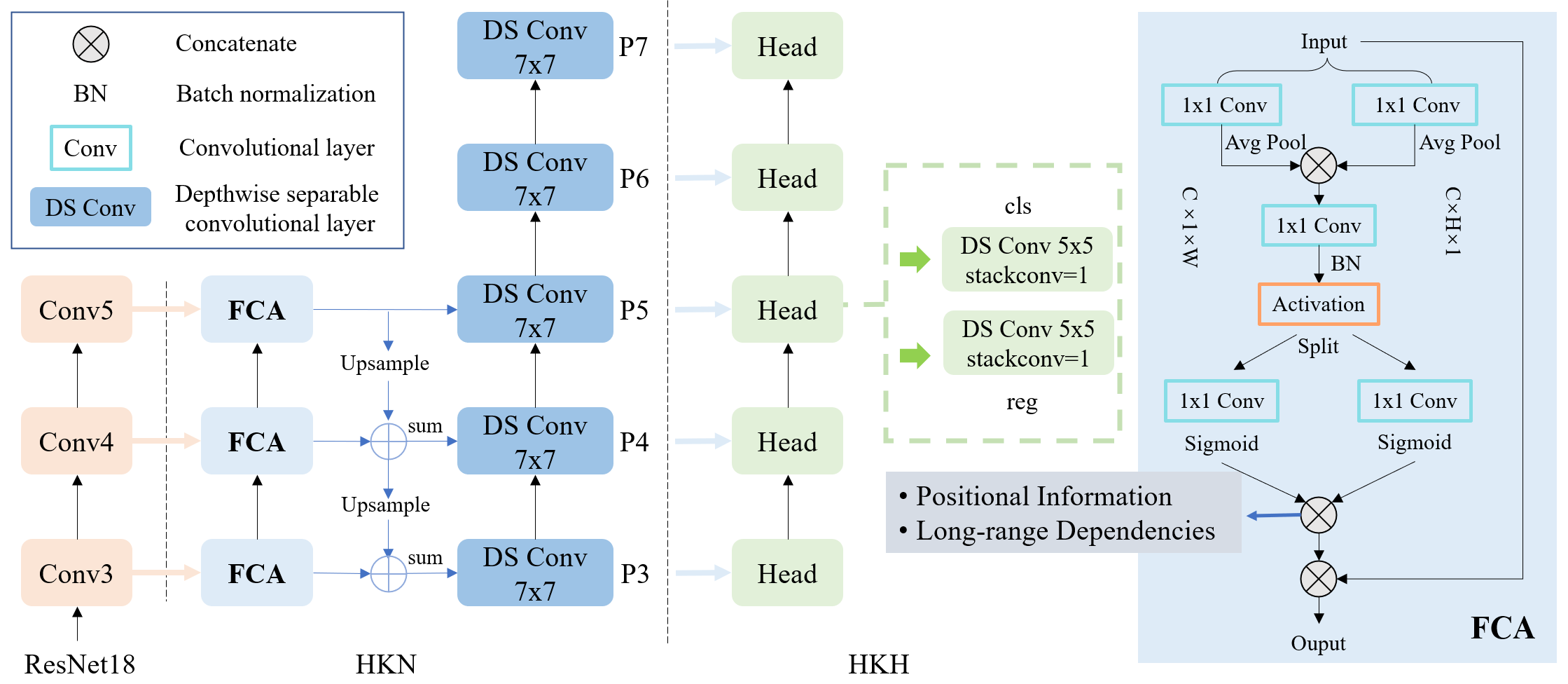}
		\caption{The detailed framework of the proposed method. This framework contains a lightweight backbone with a heterogeneous knowledge neck (\textbf{HKN}) and uses a general distribution in the shared heterogeneous knowledge head (\textbf{HKH}) to model the  uncertainty of the bounding box. The HKN includes a feature coordinate attention (\textbf{FCA}) module, which is used to model long-range dependencies and positional information between different channels.}
		\label{fig3}
	\end{center}
\end{figure*}

Therefore, we propose a heterogeneous self-distillation method that aims to extract more distillable information while ensuring that the model size meets the requirements for real-time fault detection of the freight train braking system. In Fig.~\ref{fig3}, we propose a feature coordinate attention (FCA) module for capturing spatial information between channels. The FCA obtains positional information and long-range dependencies by embedding spatial coordinate data into feature maps. 
In Section~\ref{HKH}, we model the localization information as a general distribution to obtain more reliable and accurate bounding box estimates for complex scenes. Moreover, we integrate a novel loss function that calculates three loss functions to obtain a robust student network. By maintaining the consistency of the deep features in heterogeneous teacher and student, the learning ability of the student in spatial correlation and localization information is enhanced. The proposed HSD FTI-FDet distills this abundant feature information into the student model to enhance detection performance.

\subsection{Heterogeneous Knowledge Neck}
\label{HKN}
As discussed before, student models generally adopt a lightweight backbone to extract feature maps of different scales and then employ FPN~\cite{FPN} for feature fusion. The FPN achieves a good prediction effect by fusing feature maps with different levels to realize a fusion of high-resolution and high-semantic information. Nevertheless, FPN merely focuses on semantic information, ignoring the interdependence between different channels and spatial information, losing a lot of valuable information in the fusion process.

Therefore, we propose a novel neck to capture channel and spatial information, as illustrated in Fig.~\ref{fig3}. Our design refers to three criteria: 
i) The new fusion component must be as simple and lightweight as possible in train fault detection. 
ii) The captured positional information can be fully utilized to attract more attention to the area of interest. 
iii) It should also efficiently capture the dependencies between different channels, which is significant in the existing research~\cite{SE}.

We first adjust the dimension of feature maps through 1$\times$1 convolution to integrate cross-channel information without sacrificing resolution. Then, the low-dimension feature maps are usually subjected to global average pooling to encode spatial information. However, this operation ignores the location information in the spatial structure, which leads to the bias of the area of interest in the visual detection task. To capture the ignoring positional information, we decompose the global average pooling operation along the vertical and horizontal directions. Specifically, through parallel 1D encoding operations, we employ Eq.~(\ref{FCA_1}) to compress global spatial information in both directions, and two different direction-aware feature maps are integrated as the input features.
\begin{equation} 
	\label{FCA_1}
	\left\{ \begin{array}{l}
		{z_{c}^{h} = \frac{1}{W}{\sum\limits_{0 \leq i < W}\,}x_{c}\left( {h,i} \right)} \\
		{z_{c}^{w} = \frac{1}{H}{\sum\limits_{0 \leq j < H}\,}x_{c}\left( {j,w} \right)} \\
		\end{array} \right..
\end{equation}

The input $x_{c}\left( {h,i} \right)$, $x_{c}\left( {j,w} \right)$ is directly from the 1$\times$1 convolution which can be viewed as the local descriptor at height $h$ and width $w$ direction.  We adopt pooling kernels $(H, 1)$ and $(1, W)$ to encode each channel and $z_{c}^{h}$, $z_{c}^{w}$ is the encoding result of the two direction. Different direction-aware feature maps can be interpreted as a tensor of spatial descriptors at the channel level in the corresponding direction. The statistics of these descriptors can express the whole image. Each direction-aware feature map catches the positional information of the input feature map in that direction, which helps the network concentrate on the region of interest for accurate localization.

After obtaining the precise encoding positional information within channels, we further capture long-range dependencies between channels, making our model pay more attention to regions of interest. First, we concatenate the two direction-aware feature tensors and apply 1$\times$1 convolution ($F_{1}$) to integrate the information. {Subsequently, we perform batch normalization on the feature maps and then apply a nonlinear activation function $\delta$ to guarantee the expression capability of the model. Then the intermediate feature map of the interest region can be defined as $\mathcal{G}  =  {\delta\left( {F_{1}\left( \left\lbrack {z_{c}^{h},z_{c}^{w}} \right\rbrack \right)} \right)}$, where $[ \cdot, \cdot]$ displays the concatenation operation at the spatial dimension. }

To model the long-range dependencies between different channels, {we split the intermediate feature map $\mathcal{G} $ in both directions along the spatial dimension to generate channel feature tensors  ${\mathcal{G}}^{h}$, ${\mathcal{G}}^{w}$. Another two convolutions $F_{h}$ and $F_{w}$ are conducted to separately convert ${\mathcal{G}}^{h}$, ${\mathcal{G}}^{w}$ into the same channel as the input $x$, yielding:}
\begin{equation}
	\label{FCA_2}
	\left\{ \begin{array}{l}
		{\mathbf{g}^{h} = \sigma\left( {F_{h}\left( {\mathcal{G}}^{h} \right)} \right)} \\
		{\mathbf{g}^{w} = \sigma\left( {F_{w}\left( {\mathcal{G}}^{w} \right)} \right)} \\
		\end{array} \right.,
\end{equation}
{where $\sigma$ is the $\textit{sigmoid}$ function, which is used to obtain the final attention weight $\mathbf{g}^{h}$, $\mathbf{g}^{w}$ of the input feature map in height and width. Finally, we multiply the weights with the corresponding elements of the original input feature map to get the final output feature map. The output of FCA is defined as:}
\begin{equation}
	\label{FCA_3}
	FCA(i,j) = x_{c}(i,j) \times \mathbf{g}^{h}(i) \times \mathbf{g}^{w}(j).
\end{equation}

{As previously mentioned, FCA applies attention horizontally and vertically on the input tensor concurrently. Each element in the attention maps indicates if the object of interest is present in the appropriate row or column. Consequently, this encoding procedure allows our method to locate the exact position of the fault area.} 

\begin{figure}[!t]  
	\begin{center}    
		\subfloat[]{\includegraphics[width=1.0in]{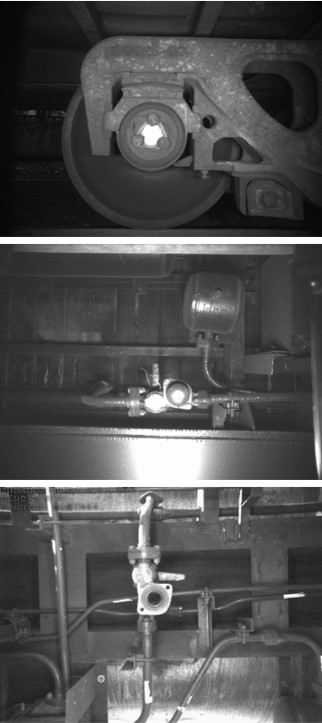}}
		\hspace{0.1em}	
		\subfloat[]{\includegraphics[width=1.0in]{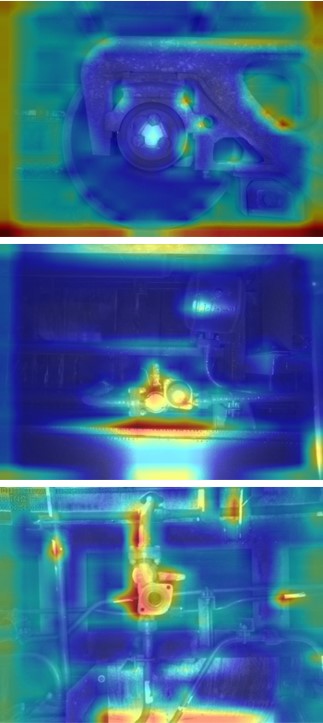}}
		\hspace{0.1em}	
		\subfloat[]{\includegraphics[width=1.0in]{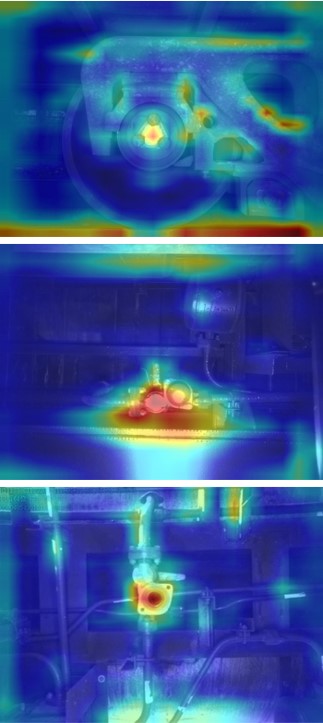}}
		\caption{Class activation map comparison of P3 in the neck. (a) Input images; (b) The class activation map of FPN; (c) The class activation map of HKN. The response area in (c) is more accurate and has fewer aliasing effects after upsampling.}
		\label{fig4}
	\end{center}
\end{figure}

In addition, considering that FCA may generate some extra information and the aliasing effect after upsampling, we utilize depthwise separable convolution (DS Conv)~\cite{Dw} to constrain this redundant information. The incorporation of DS Conv can make more effective information to participate in the following prediction of the network and enhance the precision. Moreover, DS Conv greatly reduces the computational cost and model size, demonstrating computational efficiency in object detection. 
DS Conv is defined as:
\begin{equation}
	\label{Dw}
	{\hat{G}}_{m,n,p} = {\sum\limits_{i,j}\,}{\hat{K}}_{i,j,p} \cdot F_{m + i - 1,n + j - 1,p},
\end{equation}
where $\hat{K}(\cdot)$ represents the kernel of size of DS Conv, $\hat{G}(\cdot)$ represents the output of the filtered feature map. The optimal performance of our network is attained with a convolution kernel size of 7. Intuitively, we generate the class activation map of HKN and contrast them with those of the original FPN to verify the benefits of our HKN which is shown in Fig.~\ref{fig4}. The visualization comparison shows that HKN provides more significant features and better applicability in identifying faulty areas. 

\subsection{Heterogeneous Knowledge Head}
\label{HKH}
Due to the complicated environment for train fault detection, we devote more attention to the localization quality of the bounding box in detection tasks. 
The regular operation of bounding box regression is to model the regression label as a Dirac delta distribution, which merely concentrates on the ground-truth localization and ignores the uncertainty of the edge. The Gaussian distribution represents the ambiguity by introducing variance, but capturing the real distribution of bounding boxes is difficult.

Therefore, we introduce the general distribution of the bounding box, which utilizes $\mathcal{B} = \left\{ {t,b,l,r} \right\}$ as the basic representation. General distribution models the uncertainty of the edges through a flexible probability distribution to obtain the true distribution of bounding boxes. Assuming that $\varepsilon$ is an edge of the bounding box, the prediction $\hat{\varepsilon}$ can generally be expressed as the probability distribution:
\begin{equation}
	\hat{\varepsilon} = \int_{\varepsilon_{min}}^{\varepsilon_{max}}x{P(x)}  \,dx.
\end{equation}
${P(x)}$ is the corresponding regression coordinate probability of the regression range in $\left\lbrack {\varepsilon_{min},\varepsilon_{max}} \right\rbrack$ and $x$ is the regression coordinate. 

For the discrete output of the CNN, we discretize the regression range $\left\lbrack {\varepsilon_{min},\varepsilon_{max}} \right\rbrack$ into $n$ sub-intervals $\left\lbrack {\varepsilon_{0}, \varepsilon_{1}, ..., \varepsilon_{n}} \right\rbrack$.
Meanwhile, the \textit{SoftMax} function is employed to convert the predictions of all possible positions into a probability distribution. For simplicity, the position probability on $\left\lbrack {\varepsilon_{i}, \varepsilon_{i+1}} \right\rbrack$ is denoted as $P(i)$. Given the properties of discrete distributions, the predicted value $\hat{\varepsilon}$ is expressed as:
\begin{equation}
\hat{\varepsilon} = {\sum\limits_{i = min}^{max}{P(x_{i})x_{i}}}.
\end{equation}
The general representation provides the uncertainty and true distribution of the predicted box through the flatness of the probability distribution. 

To guide the network and devote more attention to the value near the true label $\varepsilon$, we calculate the probabilities $P(i)$ and $P(i+1)$ of the two nearest positions $\varepsilon_{i}$ and $\varepsilon_{i+1}$ of label $\varepsilon$. Generalizing $\hat{\varepsilon} = P(i)\varepsilon_{i} + P(i+1)\varepsilon_{i+1}$ to be the entropy of the distribution of two positions infinitely close to the label, where 
\begin{equation}
	P(i) = \frac{\varepsilon_{i+1} - \varepsilon}{\varepsilon_{i} - \varepsilon_{i+1}}, \qquad
	P(i+1) = \frac{\varepsilon - \varepsilon_{i}}{\varepsilon_{i+1} - \varepsilon_{i}}.
\end{equation} 
By iteratively updating $P(i)$ and $P(i+1)$, the network can quickly focus on the distribution of the adjacent regions of the target localization. When the quality estimation of the sample is inaccurate and deviates from the label $\varepsilon$, we increase the penalty on the network to ensure that the estimated regression target gradually approaches the corresponding label. Subsequently, we combine the above two extended parts, which is termed as $L_{FDL}$:
\begin{equation}
	\label{FC Loss}
	L_{FDL} = \left( \varepsilon_{i+1} - \varepsilon\right)\log \left(P(i)\right) + \left( \varepsilon - \varepsilon_{i}\right)\log \left(P(i+1)\right).
\end{equation}

As discussed before, HKN embeds channel-level correlation and spatial-level coordinate information into features through different encodings. In other words, the soft target of the heterogeneous teacher model has more detailed information. Therefore, we utilize Kullback-Leibler to guarantee the consistency of the deep features of the teacher and student model during the training phase. Given the output logits $z_{T}$ and $z_{S}$ for edge $\varepsilon$, the edges by teacher model $T$ and $S$ can be represented by probability distributions ${p_{T}^{\tau}}=\theta  \left(z_{T},\tau \right)$ and ${p_{S}^{\tau}}=\theta \left( z_{S}, \tau \right)$, where $\tau$ is the distillation temperature, $\theta $ is the SoftMax function. And finally, HKD loss for all the four edges of bounding box $\mathcal{B}$ can be formulated as:
\begin{equation}
	\label{HKD_loss}
	L_{HKD}\left( {\mathcal{B}_{S},\mathcal{B}_{T}} \right) = {\sum\limits_{\varepsilon \in \mathcal{B}}{L_{KL}}^{\varepsilon }}\left({p_{S}^{\tau}},{p_{T}^{\tau}} \right).
\end{equation}

The total loss of the student model $S$ throughout the training phase is presented as:
\begin{equation}
	\label{Loss}
	Loss = L_{FCL} + {\lambda_1}L_{FDL} + {\lambda_2}L_{FRL} + L_{HKD}.
\end{equation}
where $L_{FCL}$ adopts the binary cross entropy loss to denote the classification loss between prediction with ground-truth category label. $L_{FDL}$ mainly acts on the probability distribution, boosting the efficiency of learning a more reasonable probability distribution. {We also utilize GIoU loss~\cite{GIoU} as fault classification loss $L_{FRL}$ to further adjust the position of the prediction to improve the accuracy. In addition, $L_{HKD}$  utilizes KL divergence to transfer abundant feature knowledge from heterogeneous teachers to the student model, further strengthening the performance of the lightweight detection model. As for the hyper-parameters, $\lambda_1$ and $\lambda_2$ are set to 0.25 and 2, respectively. Specifically, we assign different weights enabling our method to balance the trade-off between different losses and maximize the overall performance of the model.} HSD FTI-FDet can be applied to a diverse range of detectors and is not merely constrained between one-stage and two-stage detection methods.

\begin{figure}[!t]
	\centering
	\includegraphics[width=3.3in]{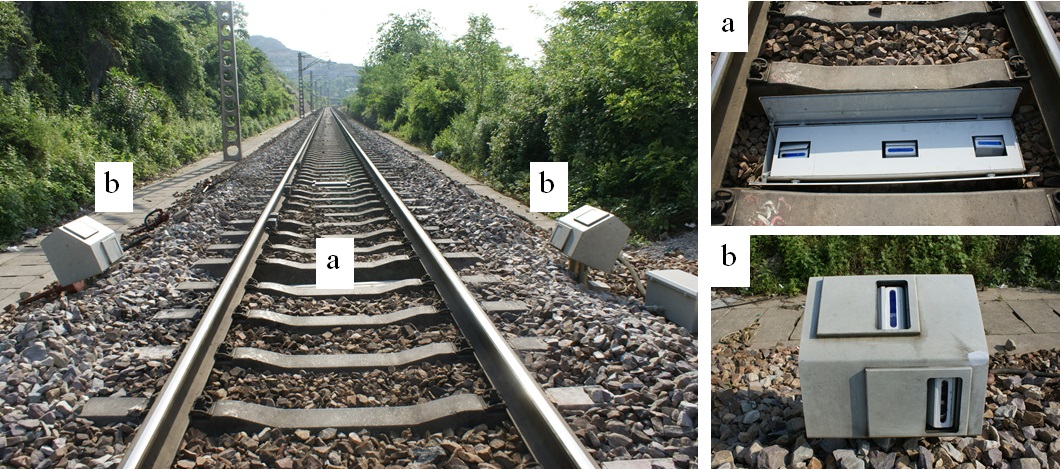}
	\caption{Vision-based efficient fault detection of freight trains based on heterogeneous self distillation method. High-speed cameras and auxiliary lights are positioned on both edges and the center of railway tracks as image-collecting devices.}
	\label{fig5}
\end{figure}

%% file: files/4-Experiments.tex

\section{Experiments and Analysis}
\label{sec:Experiments}
In this part, we describe four fault datasets~\cite{Zhang_TIM} of the freight train braking systems, then employ seven evaluation metrics to validate the effectiveness of the proposed method. Moreover, we analyze the overall framework, the efficiency of heterogeneous knowledge neck (HKN), heterogeneous knowledge head  (HKH), and various modules. Finally, we compare our HSD FTI-FDet with state-of-the-art detection methods on the four datasets.

\subsection{Experimental Setup}

\subsubsection{Datasets}
To validate the effectiveness of our method, we use the labeling tool to create four datasets~\cite{Zhang_TIM} for the experiments, including the bogie block key, cut-out cock, dust collector, and fastening bolts on the brake beam. These four elements are essential to guarantee security for the freight train braking systems.

\begin{itemize}
\item {\textbf{Bogie Block Key} (BBK): The bogie block key is set between the train wheel axle with the bogie to avoid vibration and displacement caused by the separation between them during normal operation. This dataset is divided into 5440 and 2897 images, which are prepared for training and testing, respectively.}
\item {\textbf{Cut-out Cock} (COH): Cut-out Cock is a crucial element for turning off airflow from the main fuel tank to the brake pipe, and its function is to cut off the switch of the brake pipe. This dataset is divided into 815 and 850 images, which are prepared for training and testing, respectively.}
\item {\textbf{Dust Collector}{ (DCM): Dust Collector is installed on the brake pipe to collect the impurities in the pressure air to avoid abnormal wear and even failure of the brake device caused by the impurities in the power source. }This dataset is divided into 815 and 850 images, which are prepared for training and testing, respectively.}
\item {\textbf{Fastening Bolt (Missing) on Brake Beam} (FBM): During the braking process of the train, the enormous braking force probably destroys the fastening bolts, even directly falling off. Then the brake beam fails and causes the train to decelerate, which poses a huge safety hazard. This dataset is divided into 1724 and 1902 images, which are prepared for training and testing, respectively.}
\end{itemize}

Nevertheless, as shown in Fig.~\ref{fig5}, the image acquisition equipment is usually positioned in the wild. Many other parts at the bottom of the train may block these critical components, making it challenging to accurately locate the fault area and increasing the difficulty of train fault detection.

\subsubsection{Evaluation Metrics}
{We use seven indexes: correct detection rate (CDR), missing detection rate (MDR), false detection rate (FDR), training memory usage, testing memory usage, inference speed, and model size to verify the performance of detectors. The CDR, MDR, and FDR are utilized to evaluate the accuracy and defined as follows: for a dataset consisting of $m$ fault images and $n$ normal images, after the operation of the detector, $a$ images are classified as fault images, while $c$ images are classified as normal images. Among them, $b$ images in $a$ and $d$ images in $c$ are detected by mistakes, respectively. Then the index can be defined as:  $MDR = b/(m+n), FDR = d/(m+n), CDR = 1-FDR-MDR.$ Since the limitation of application scenarios, the fault detector with low computational costs and fast detection speed is more acceptable. The memory usage of the model during training and testing is also essential, which reflects the dependence of the detector on the hardware system.
In addition, mCDR, mMDR, and mFDR are computed as the mean value of CDRs, MDRs, and FDRs to evaluate the accuracy and robustness of fault detectors on four fault datasets.}

 \begin{table*}[!t]
	\footnotesize
	\renewcommand\arraystretch{1.3}
	\caption{Comparative experiments between standard distillation method with self distillation method on various backbones of ResNet18(\textbf{R18}), ResNet50(\textbf{R50}), ResNet101(\textbf{R101}), MobileNetV3(\textbf{MV2}), ShuffleNetV2(\textbf{SV3}). {Schedule (1x): single-scale training 12 epochs.} }
	\label{backbone}
	\centering
	\begin{tabular}{cc|c|ccc|c}
	\hline
		\multirow{2}[2]{*}{Teacher}                     &
		\multirow{2}[2]{*}{Student}                     &
		\multirow{2}[2]{*}{Schedule}                     &
		\multirow{2}[2]{*}{\begin{tabular}[c]{@{}c@{}}CDR(\%)$\uparrow$  \end{tabular}}   &
		\multirow{2}[2]{*}{\begin{tabular}[c]{@{}c@{}}FDR(\%)$\downarrow$  \end{tabular}}   &
		\multirow{2}[2]{*}{\begin{tabular}[c]{@{}c@{}}MDR(\%)$\downarrow$  \end{tabular}}   &
		\multirow{2}[2]{*}{\begin{tabular}[c]{@{}c@{}}Model Size (MB)\end{tabular}} \\
		  &       &       &       &      &  \\
	\hline
		R101& R18 	& 1x	& 99.48  & 0.04  & 0.48  & 333 \\
		R101& R34 	& 1x	& 99.41  & 0.24  & 0.35 & 419 \\
		R101& R50 	& 1x	& 98.24  & 0.79  & 0.97  & 441 \\
		R101& MV3 	& 1x	& 95.17  & 2.28  & 2.55  & 252 \\
		R101& SV2 	& 1x	& 73.08  & 14.73  & 12.19  & 252 \\
	\hline
	\hline
		R18 & R18 	& 1x	& 98.96  & 0.49  & 0.55  & 219 \\
		MV3 & MV3 	& 1x	& 91.44  & 3.73  & 4.83  & 99 \\
		SV2 & SV2 	& 1x 	& 77.05  & 7.90  & 15.05  & 98 \\
	\hline
	\end{tabular}
\end{table*}%

 \begin{table*}[!t]
	\footnotesize
	\renewcommand\arraystretch{1.3}
	\caption{Ablation study of different module designs in HKN on BBK dataset with ResNet18.}
	\label{Neck}
	\centering
   \begin{tabular}{l|ccc|c}
	\hline
		\multirow{2}[2]{*}{Module}                     &
		\multirow{2}[2]{*}{\begin{tabular}[c]{@{}c@{}}CDR(\%)$\uparrow$  \end{tabular}}   &
		\multirow{2}[2]{*}{\begin{tabular}[c]{@{}c@{}}FDR(\%)$\downarrow$  \end{tabular}}   &
		\multirow{2}[2]{*}{\begin{tabular}[c]{@{}c@{}}MDR(\%)$\downarrow$  \end{tabular}}   &
		\multirow{2}[2]{*}{\begin{tabular}[c]{@{}c@{}}Model Size(MB)\end{tabular}} \\
				  &       &       &       &      \\
		\hline
		P3-P5 & 82.26  & 9.08  & 8.66  & 210 \\
		P3-P6 & 80.81  & 13.15  & 6.04 & 213 \\
		P3-P7 & 98.96  & 0.49  & 0.55  & 219 \\
		\hline
		\hline
		\quad+DS conv3x3 & 99.21  & 0.38  & 0.41  & 189 \\
		\quad+DS conv5x5 & 99.21  & 0.24  & 0.55  & 190 \\
		\quad+DS conv7x7 & 99.28  & 0.48  & 0.24  & 190 \\
		\hline
		\hline
		\qquad+FCA    & 98.52  & 0.38  & 1.10  & 220 \\
		\qquad+FCA+DS conv7x7 & 99.34  & 0.31  & 0.35  & 190 \\
		\hline
		\end{tabular}
\end{table*}%

\subsubsection{Implementation Details}
Specifically, all models are trained with back-propagation and SGD optimizer with 0.9 momentum and $10^{-5}$ weight decay. Experiments are conducted with batch size 4 and 1$\times$ (12 epochs) training schedule for all ablation studies. The learning rate decay is divided by $10 \times$ at the 8th and 11th epochs with $10^{-4}$ initial learning rate. The main experiments are confirmed on four fault datasets.

Additionally, the input image is scaled to 700$\times$512 and initialized with mean and variance. The probability that the image is randomly flipped left and right is 0.5. Our experiments are carried out on a single GTX2080Ti GPU under Ubuntu 20.04 sever. Besides, the hyperparameters are fine-tuned during the training phase to achieve the greatest results about each model in Table~\ref{SOTA}.

\subsection{Ablation Study}
\label{sec:ablation}
\subsubsection{Distillation Strategy}
We first explore the overall structure to select a more appropriate framework for fault detection. We experiment with different representative network architectures using KD and self-distillation, utilizing the bogie block key datasets. We employ the identical training configuration following \cite{LD_2021}, except for linearly scaling up the initial learning rate and batch size. As shown in Table~\ref{backbone}, KD generally achieves higher accuracy than self-distillation. For the model of teacher ResNet101, student ResNet18 achieves 99.48$\%$, while the model size is 333 MB which cannot meet the actual requirements of the train detection environment. So we explore the self-distillation method and found that despite the loss in accuracy, the model size was reduced by nearly a third. In the self-distillation experiment, compared with ShuffleNetV2~\cite{ShufflenetV2}, MobileNetV3~\cite{MobilenetV3} has a model size of 219 MB, but the CDR index can reach 98.96$\%$, which is more suitable for scenarios with high accuracy requirements.

\begin{figure}[!t]  
	\begin{center}    
		\subfloat[]{\includegraphics[width=1.0in]{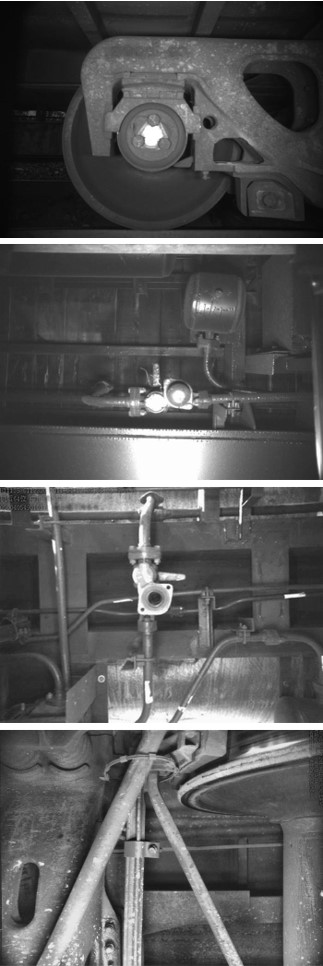}}
		\hspace{0.1em}	
		\subfloat[]{\includegraphics[width=1.0in]{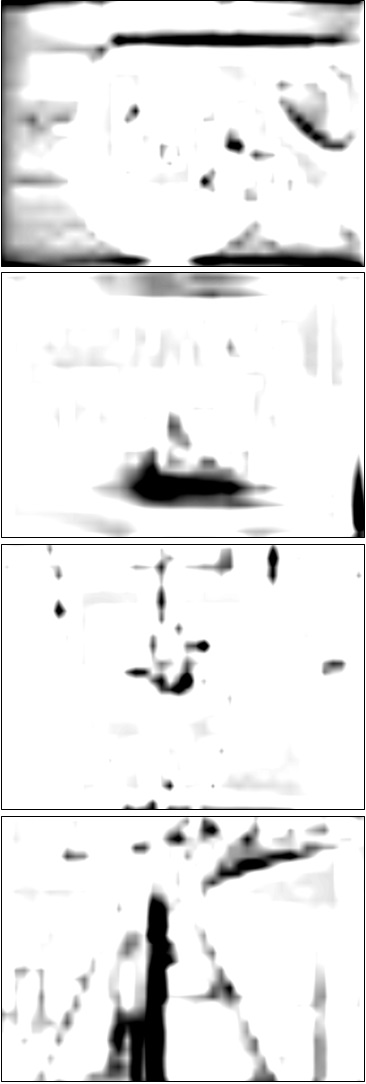}}
		\hspace{0.1em}	
		\subfloat[]{\includegraphics[width=1.0in]{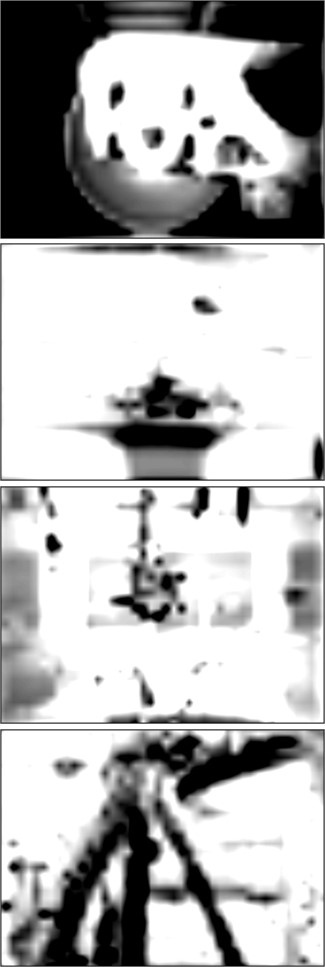}}
		\caption{Visualization comparison of the average feature maps. (a) Input images; (b) The feature map from Conv3 in the teacher model; (c) The feature map from Conv3 in the student model. (c) It acquires more edge localization information, showing that the distillation is effective.}
		\label{fig6}
	\end{center}
\end{figure}

\begin{table*}[!t]
	\footnotesize
	\renewcommand\arraystretch{1.3}
	\centering
	\caption{Ablation study of different component analysis (HKN $\&$ HKH) in overall framework on BBK dataset.}
	\label{baseline}
		\begin{tabular}{cc|c|cc|ccc|c}
			\hline
			\multirow{2}[2]{*}{Teacher}                     &
			\multirow{2}[2]{*}{Student}                     &
			\multirow{2}[2]{*}{Schedule}                     &
			\multirow{2}[2]{*}{HKN}                     &
			\multirow{2}[2]{*}{HKH}                     &
			\multirow{2}[2]{*}{\begin{tabular}[c]{@{}c@{}}CDR(\%)$\uparrow$  \end{tabular}}   &
			\multirow{2}[2]{*}{\begin{tabular}[c]{@{}c@{}}FDR(\%)$\downarrow$  \end{tabular}}   &
			\multirow{2}[2]{*}{\begin{tabular}[c]{@{}c@{}}MDR(\%)$\downarrow$  \end{tabular}}   &
			\multirow{2}[2]{*}{\begin{tabular}[c]{@{}c@{}}Model Size(MB)\end{tabular}} \\
			&	&	&	&       &       &       &      &		\\
			\hline
			R101& R18 	& 1x	& -- 				& --       			& 99.48  & 0.04  & 0.48  & 333  \\
			R18 & R18 	& 1x	& -- 				& --       			& 98.96  & 0.49  & 0.55  & 219 \\
			R18 & R18 	& 1x	&$\checkmark$    	& --      			& 99.34  & 0.31  & 0.35  & 190 \\
			R18 & R18 	& 1x	&--					& $\checkmark$     	& 99.90  & 0.03  & 0.07  & 170 \\
			R18 & R18 	& 1x	&$\checkmark$    	& $\checkmark$     	& 99.55  & 0.28  & 0.17  & 96 \\
			\hline
		\end{tabular}
\end{table*}

\begin{figure}[t]
	\centering
	\includegraphics[width=1.5in]{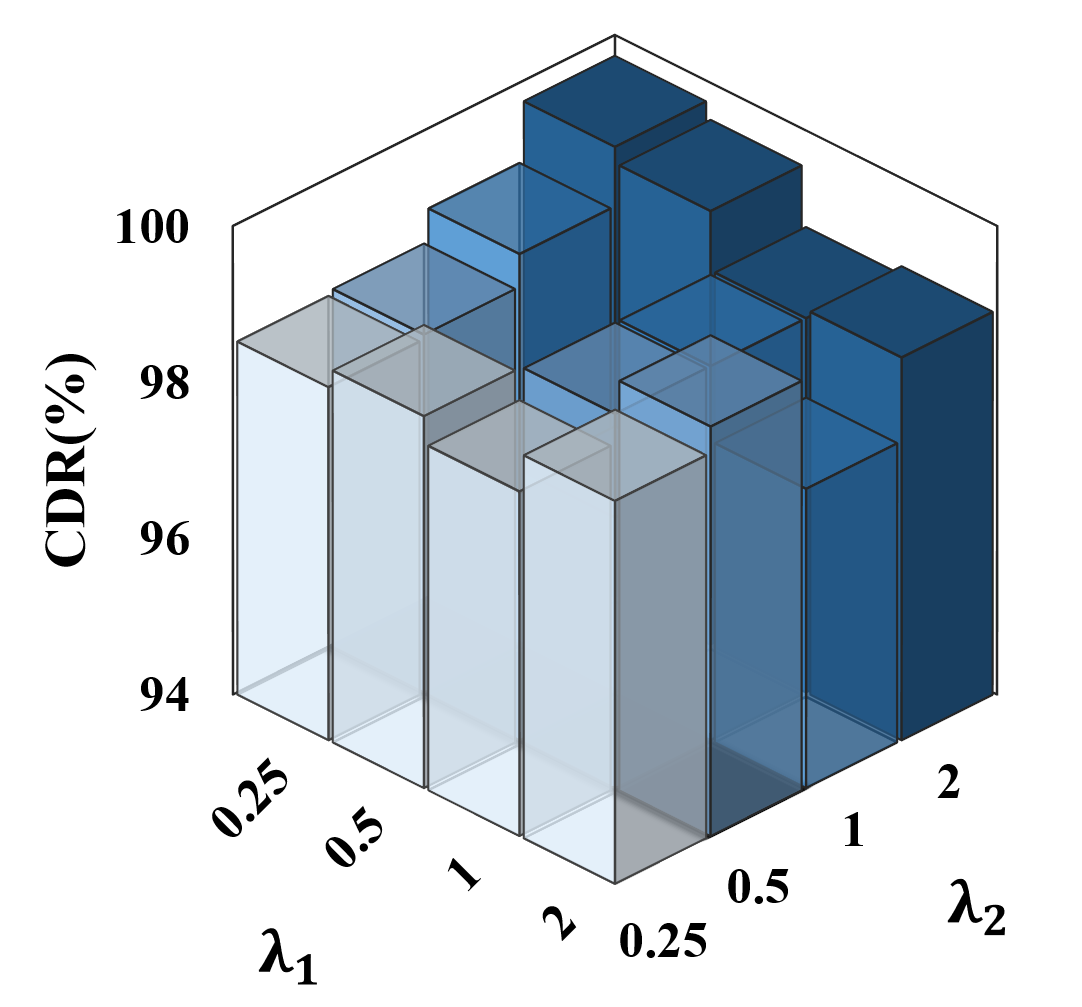}
	\caption{{Comparison of the experimental results of the hyperparameter settings of $L_{FDL}$ and $L_{FRL}$. We set $\lambda_1$=0.25 and $\lambda_2$=2 by default. Among them, $L_{FDL}$ is used to improve the efficiency of learning a more reasonable probability distribution; $L_{FRL}$ is used to adjust the predicted position further and improve the accuracy. }}
	\label{fig_loss}
\end{figure}

\begin{figure}[!t]
	\centering
	\includegraphics[width=1.6in]{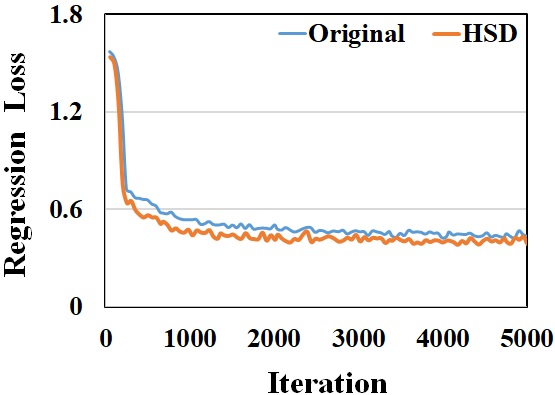}
	\includegraphics[width=1.6in]{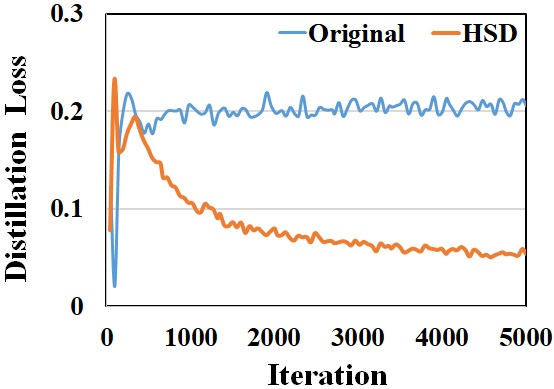}
	\caption{The comparison of regression and distillation loss between the baseline~\cite{LD_2021} and HSD FTI-FDet during training iterations when the student is ResNet18. The regression curve converges faster, and the oscillation amplitude of the distillation loss is alleviated, indicating that our method is easier to obtain the global optimum.}
	\label{fig7}
\end{figure}

\begin{table}[!t]
	\footnotesize
	\renewcommand\arraystretch{1.3}
	\begin{center}
	\caption{Temperature \textbf{$\tau$} in HSD on BBK dataset: The generalized Softmax function with appropriate \textbf{$\tau$} brings  significant benefits. We set $\textbf{$\tau$}=15$ by default. }
	\label{T}
		\begin{tabular}{p{12mm}<{\centering}p{12mm}<{\centering}p{12mm}<{\centering}p{12mm}<{\centering}cccc}
			\toprule
			\multirow{2}[2]{*}{$\tau$}                     &
			\multirow{2}[2]{*}{\begin{tabular}[c]{@{}c@{}}CDR(\%)$\uparrow$  \end{tabular}}   &
			\multirow{2}[2]{*}{\begin{tabular}[c]{@{}c@{}}FDR(\%)$\downarrow$  \end{tabular}}   &
			\multirow{2}[2]{*}{\begin{tabular}[c]{@{}c@{}}MDR(\%)$\downarrow$  \end{tabular}}   \\
				&       &       &    \\
			\midrule
			1     & 98.69 	& 0.65  & 0.66  	\\
			5     & 97.89 	& 1.49  & 0.62  	\\
			10    & 99.55 	& 0.28  & 0.17  	\\
			15    & 99.76	& 0.14  & 0.10   	\\
			20    & 98.76	& 0.76  & 0.48 	\\ \bottomrule
		\end{tabular}
	\end{center}
\end{table}

\subsubsection{Framework}
To explore our final structure, we first investigate the performance of HKN. In Table~\ref{Neck}, we perform experiments for the layers $P3$ $\sim$ $P5$, $P3$ $\sim$ $P6$ and $P3$ $\sim$ $P7$ of output channels separately, and we find that the $P3$ $\sim$ $P7$ perform better on the dataset, with an increase of nearly 18$\%$ compared to other numbers of CDRs. This indicates that the output channels of the five layers capture more useful information. On this basis, we use DS Conv to remove the aliasing phenomenon of upsampling. At the same time, we conducted ablation experiments on different sizes of convolution kernels. The CDR index can achieve 99.28$\%$, and the model size decreases by 29M while the convolution kernel size is 7. In the last column of Table~\ref{Neck}, we can see that if FCA is used alone, the accuracy drops from 98.96$\%$ to 98.52$\%$. This is because FCA causes more invalid information to enter the channel when focusing on positional information and long-distance information, while DS Conv is better at eliminating the aliasing effect, which can effectively remove this redundant information. It can be seen from Table ~\ref{Neck} that when DS Conv and FCA are used in combination, the CDR index is the highest, indicating that our HKN is more suitable for detecting the fault in train images.


\begin{table*}
	\centering
	\renewcommand\arraystretch{1.5}	
	\caption{{Detailed comparison with the state-of-the-art methods on four typical fault datasets. \textbf{T}: Teacher. \textbf{S}: Student}}
	\label{SOTA}
	\footnotesize
	\begin{tabular}{llp{6mm}<{\centering}p{6mm}<{\centering}p{6mm}<{\centering}p{8mm}<{\centering}p{8mm}<{\centering}p{10mm}<{\centering}p{8mm}<{\centering}ccccccc}
		\toprule
		\multirow{2}[2]{*}{Methods}            &
		\multirow{2}[2]{*}{\hspace{3.5em}Backbone}           &
		\multirow{2}[2]{*}{\begin{tabular}[c]{@{}c@{}}mCDR\\      (\%)$\uparrow$\end{tabular}} &
		\multirow{2}[2]{*}{\begin{tabular}[c]{@{}c@{}}mMDR\\      (\%)$\downarrow$\end{tabular}} &
		\multirow{2}[2]{*}{\begin{tabular}[c]{@{}c@{}}mFDR\\      (\%)$\downarrow$\end{tabular}} &
		\multicolumn{2}{c}{Memory(MB)}         &
		\multirow{2}[2]{*}{\begin{tabular}[c]{@{}c@{}}Inference\\      (s/image)\end{tabular}} &
		\multirow{2}[2]{*}{\begin{tabular}[c]{@{}c@{}}Model\\      size(MB)\end{tabular}}   \\ \cmidrule(r){6-7}
												&                  &       &      &       & Train** & Test &       &        \\
		\midrule
		\multicolumn{9}{c}{\textbf{Fault Detectors}}                                                                    \\ \midrule
		HOG+Adaboost+SVM \cite{Zhang_TIM}      & \hspace{5.2em}--               & 94.39 & 1.35 & 4.26  & --    & --   & 0.049 & --   \\
		FAMRF+EHF \cite{Sun}                   & \hspace{5.2em}--               & 94.96 & 1.00 & 4.04  & --    & --   & 0.725 & --   \\
		Cascade detector (LBP) \cite{Sun}      & \hspace{5.2em}--               & 88.69 & 3.77 & 7.54  & --    & --   & 0.048 & --     \\
		FTI-FDet*\cite{Zhang_TIM} & \hspace{3.9em}VGG-16 & 99.27 & 0.52 & 0.21 & -- & 1823 & 0.071 & 557  \\
		Light FTI-FDet* \cite{Zhang_TIM}        & \hspace{3.9em}VGG-16            & 98.91 & 0.49 & 0.61  & --    & 1533 & 0.058 & 90   \\
		\midrule
		\multicolumn{9}{c}{\textbf{General Object Detectors}}                                                                 \\ \midrule
		Cascade R-CNN \cite{Cascade_R-CNN}     & \hspace{3.7em}ResNet-50        & 99.33 & 0.20 & 0.47  & 1506  & 1841 & 0.061 & 553  \\
		Faster R-CNN \cite{Faster_R-CNN}       & \hspace{3.7em}ResNet-50        & 98.54 & 0.43 & 1.03  & 1453  & 1545 & 0.055 & 330  \\
		Dynamic R-CNN \cite{Dynamic_R-CNN}     & \hspace{3.7em}ResNet-50        & 98.47 & 0.84 & 0.69  & 1708  & 1427 & 0.056 & 330  \\
		Grid R-CNN \cite{Grid_R-CNN}           & \hspace{3.7em}ResNet-50        & 98.13 & 0.99 & 0.88  & 1627  & 1523 & 0.064 & 515  \\
		AutoAssign \cite{AutoAssign}    & \hspace{3.7em}ResNet-50        & 96.95 & 1.97 & 1.08  & 1528  & 1275 & 0.075      & 289  \\
		CentripetalNet \cite{CentripetalNet}   & \hspace{2.3em}HourglassNet-104 & 95.34 & 3.03 & 1.63  & 3878  & 3951 & 0.297 & 2469 \\
		Cornernet \cite{CornerNet}             & \hspace{2.3em}HourglassNet-104 & 96.42 & 1.26 & 2.32  & 3643  & 3543 & 0.245 & 2412 \\
		FCOS \cite{FCOS}                       & \hspace{3.7em}ResNet-50        & 97.53 & 1.99 & 0.48  & 1311  & 953  & 0.050 & 256  \\
		YOLOX \cite{YOLOX}                     & \hspace{2.4em}Modified CSP v5  & 96.64 & 0.49 & 2.87  & 1932  & 1047 & 0.026 & 651  \\
		FoveaBox \cite{FoveaBox}        & \hspace{3.7em}ResNet-50        & 96.82 & 2.24 & 0.94  & 1186  & 987  & 0.069      & 289  \\
		Libra R-CNN \cite{Balanced}            & \hspace{3.7em}ResNet-50        & 96.53 & 2.66 & 0.81  & 1614  & 1573 & 0.063 & 332  \\
		CenterNet \cite{Object}                & \hspace{3.7em}ResNet-50        & 98.04 & 0.66 & 1.29  & 728    & 1731 & 0.039  & 230  \\
		Deformable DETR \cite{Deformable_DETR} & \hspace{3.7em}ResNet-50        & 84.28 & 2.51 & 13.21 & 2009  & 1131 & 0.043 & 482  \\
		Sparse R-CNN \cite{Sparse_R-CNN}       & \hspace{3.7em}ResNet-50        & 90.08 & 0.38 & 9.54  & 1538  & 1951 & 0.058 & 1273 \\
		\midrule
		\multicolumn{9}{c}{\textbf{Distillation-based  Detectors$\dagger$}}                                                                   \\ \midrule
		Lad \cite{Lad}                         & T: ResNet-101\quad S:ResNet-50      & 97.67  & \textbf{0.02}  & 2.31  & 1098   	& 1631  	& 0.085 	& 389   \\
		ReviewKD \cite{ReviewKD}               & T: ResNet-101\quad S:ResNet-18        & 91.67  & 7.52  & 0.81  & 808  	& 1497  	& \textbf{0.019} 	& 471   \\
		ICD \cite{ICD}                         & T: ResNet-101\quad S:ResNet-50        & 97.20  & 1.58  & 1.22  & 972      & 1621      & 0.034     & 278   \\
		LD \cite{LD_2021}                      & T: ResNet-101\quad S:ResNet-18        & 98.43  & 0.40  & 1.17  & 666  	& 1595  	& 0.027 	& 333   \\
  	LGD $\ddagger$\cite{LGD}                         & T: ResNet-101\quad S:ResNet-101        & 98.63  & 0.60  & 0.77  & 1226     & 1287      & 0.052     & 499  \\ \bottomrule
		HSD FTI-FDet $\ddagger$                          & T: ResNet-18\hspace{1.18em} S:ResNet-18        & \textbf{98.88}      & 0.37     & \textbf{0.75}      & \textbf{581}  & \textbf{1159}  & 0.027      & \textbf{96}  \\
		\bottomrule
	\end{tabular}
	\begin{tablenotes}\scriptsize 
		\item *This approach was tested on a GTX2080Ti with publicly available Caffe~\cite{caffe}.
		\item **We fine-tune the relevant parameters of every detector to obtain the greatest performance. Considering that batch size directly impacts memory usage during the training phase, we calculate the memory usage by each batch for fairness.
		\item $\dagger$ We only record the performance of the student model in the comparison of distillation methods, and the teacher model does not participate in the comparison.
            \item $\ddagger$ These methods are self-distillation detectors.
	\end{tablenotes}
\end{table*}%

Second, we investigate the effectiveness of HKH. For the role of HKN and HKH, we report the relevant ablation study results in Table~\ref{baseline}. The results demonstrate that conducting HKN and HKH and their combination can improve the accuracy of the model by +0.38$\%$, +0.94$\%$, and +0.59$\%$ CDR, respectively. This indicates that our framework is indeed effective. The accuracy of train fault detection is increased when HKN is used alone, but it is insufficient for the high-precision scenario. When HKH is used alone, the accuracy reaches 99.9$\%$, but the model size is still not friendly enough for resource-constrained detection environments. The baseline CDR score is improved by 0.59$\%$, and the model size is decreased by 78M when the two are used together.

\subsubsection{Loss function}

Except for the GFocal, localization quality estimates, and classification scores are typically trained separately. However, they are combined during inference, leading to a gap between training and inference. To ensure the efficiency of the detector, we inherit the idea of GFocal using a joint representation to bridge the gap. 
{The optimized loss function utilizes $L_{FDL}$ and $L_{FRL}$ to calculate the loss of the object localization, which has two parameters $\lambda_1$ and $\lambda_2$ to balance the weight.  As shown in Fig.~\ref{fig_loss}, different parameters make mCDR produce various performance gains, which further illustrates that the optimized loss function in the detection head can enhance the model to obtain abundant localization information.
In Fig.~\ref{fig7}, our box regression curve converges faster, making it easier to obtain the global optimum on the train fault dataset. Compared with the baseline, our HSD FTI-FDet alleviates the oscillation amplitude of distillation loss to a certain extent which verifies the effectiveness of our loss function.}

\subsubsection{Sensitivity study of temperature $\tau$}
In Eq.~\ref{HKD_loss}, we utilize the temperature hyperparameter $\tau$ to rectify the transmission of long-range dependencies and positional information between the feature map. As shown in Table~\ref{T}, the student gains 0.21$\%$ mCDR improvement at $\tau=15$ compared with $\tau=10$, which means distillation is significantly effective. The long-range dependencies and positional information are emphasized more during the training phase, allowing the student to concentrate on crucial features more and perform better. Besides, the worst result is only 1.66$\%$ mCDR lower than the best, demonstrating that our method is not sensitive to the $\tau$.

\begin{figure}[t]  
	\begin{center}    
		\subfloat[]{\includegraphics[width=1.0in]{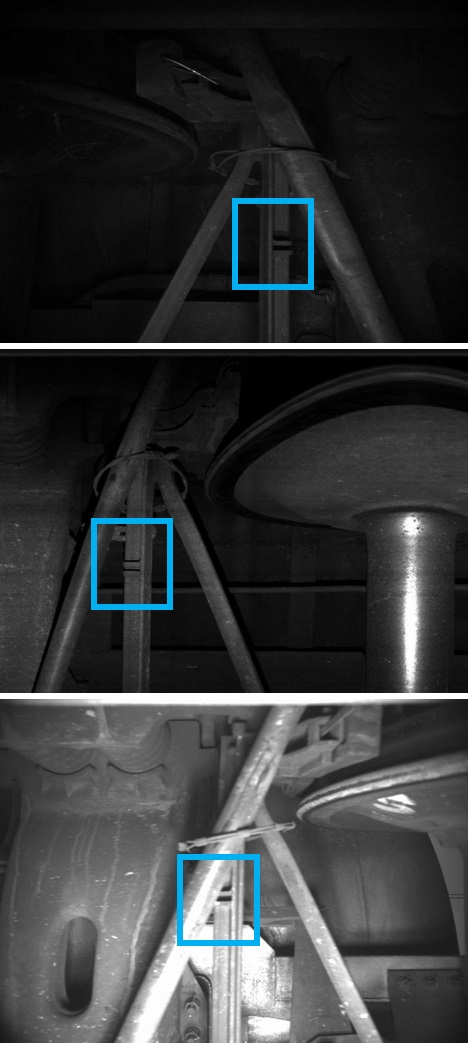}}
		\hspace{0.1em}	
		\subfloat[]{\includegraphics[width=1.0in]{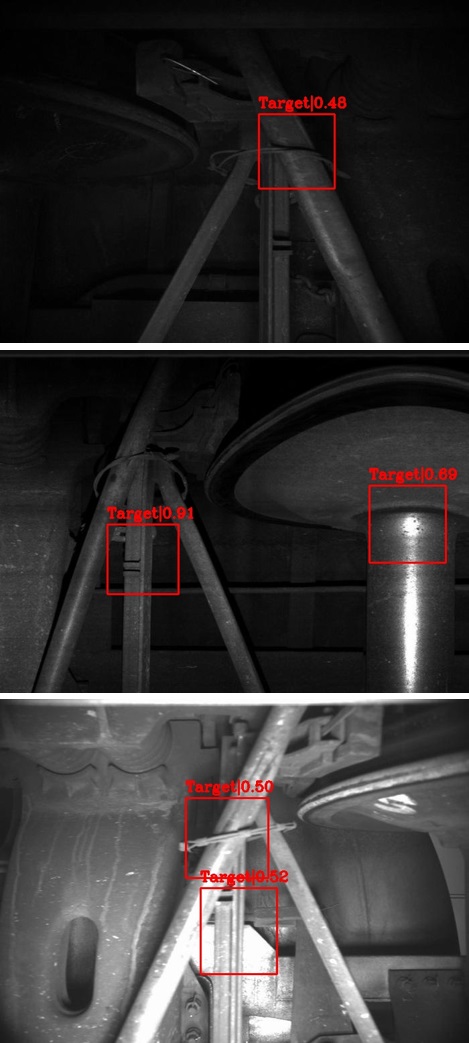}}
		\hspace{0.1em}	
		\subfloat[]{\includegraphics[width=1.0in]{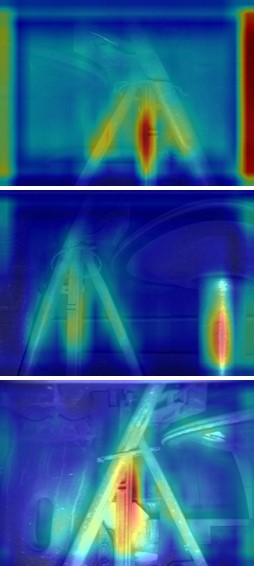}}
		\caption{{Instance analysis of abnormal scenarios on the FBM dataset. (a) Visualization of ground-truth; (b) Visualization of failure examples; (c) Class activation map. Due to the lack of such images in datasets, our method is less robust to scenes with uneven lighting in (b) such as overexposed and underexposed scenes but still realizes accurate location in the interest area in (c).  We will further expand our datasets to address this problem in the future.}}
		\label{fig8}
	\end{center}
\end{figure}

\subsection{Comparison with the state-of-the-art methods}
We compare the performance of HSD FTI-FDet with state-of-the-art detectors to further illustrate the superiority of our framework. These methods can be divided into traditional fault detectors, general object detectors, and distillation-based detectors. 

\textbf{Accuracy and model size.}
{As shown in Table~\ref{SOTA}, the mCDR of HSD FTI-FDet achieves 98.88$\%$, which is higher than all fault detectors and general object detectors except for FTI-FDet, Light FTI-FDet, and Cascade R-CNN. FTI-FDet is quite higher than ours in accuracy, but our model size is much smaller. The mCDR of Light FTI-FDet is higher than our method by 0.03$\%$, but the inference speed is only half of our HSD FTI-FDet. 
Cascade R-CNN utilizes a two-stage approach for detection and achieves 99.33$\%$ accuracy, but the model size is almost 4.5$\times$ larger than ours which can not meet the hardware resource requirement. Compared with other distillation-based detectors, HSD FTI-FDet achieves the highest detection accuracy (98.88$\%$) and the smallest model size (only 96MB). LD utilizes ResNet18 as the backbone but the model size (333MB) is 3.4$\times$ larger than ours and the mCDR is also 0.45$\%$ lower than ours, which further demonstrates the effectiveness of our distillation method. The results show that heterogeneous self-distillation combined with FCA and optimized loss function allows for large performance gains without significantly increasing the model size. 
However, since the lack of such images in datasets, our method is not satisfactory enough under different lighting conditions in Fig.~\ref{fig8}. The visualization result demonstrates that our images accurately represent the regions of interest in the model, even under varying lighting conditions, thereby providing further evidence of the robustness of our model.} We will address this by expanding the dataset by employing data augmentation in data preprocessing which simultaneously enhances the generalization ability of the model.

\textbf{Computational cost.}
Generally, speed and memory usage represent the computational cost of the model. Due to HKN adopting the efficient DS Conv, our memory usage is only 581MB during the training phase and 1159MB in the testing phase. HSD FTI-FDet achieves the smallest training memory usage in Table~\ref{SOTA}, while the general detectors are mostly more than 1000MB or even larger. 
In terms of inference speed, HSD FTI-FDet infers an image of just 0.027s ($>$ 37 FPS), faster than all traditional and general object detectors. Moreover, HSD FTI-FDet implements the same testing speed as YOLOX. Compared to FTI-FDet and Light FTI-FDet, HSD FTI-FDet is 2.6/2.2$\times$ faster. 

According to experiment results, our strategy balances resources and accuracy better than other methods and fulfills the requirements of real-time train fault detection. Experiments on four fault datasets also reveal the robustness and generality of our method for various types of faults. Therefore, HSD FTI-FDet is better suited for fault detection of freight train braking systems with environmental constraints and high accuracy requirements.

%% file: files/5-Conclusion.tex
\section{Conclusion}
\label{Conclusion}

This paper presents a lightweight and accurate framework HSD FTI-FDet based on the self-distillation method for fault detection of freight train braking systems. The proposed HSD FTI-FDet comprises a heterogeneous knowledge neck that captures long-range dependencies across channels and a heterogeneous knowledge head with a novel loss function. Experiments on four fault datasets demonstrate that HSD FTI-FDet achieves more excellent performance than other detectors. HSD FTI-FDet implements competitive accuracy 98.88$\%$ under limited hardware resources and 5.8$\times$ smaller model size than FTI-FDet. Meanwhile, our method has the same testing speed as YOLOX, 2.6$\times$ faster than FTI-FDet and 2.2$\times$ faster than Light FTI-FDet with lower memory usage.

We will concatenate on improving our method to realize real-time multi-fault detection on embedded platforms (e.g. Raspberry Pi and Jetson Nano) in the future. Additionally, we intend to employ multiple image processing methods to enhance the robustness of our method in lighting conditions and further improve detection accuracy.
\\